\documentclass[10pt,twocolumn,letterpaper]{article}

\usepackage{cvpr}
\usepackage{times}
\usepackage{epsfig}
\usepackage{graphicx}
\usepackage{amsmath}
\usepackage{amssymb}

\usepackage{verbatim}
\usepackage{multirow}

\usepackage[pagebackref=true,breaklinks=true,letterpaper=true,colorlinks,bookmarks=false]{hyperref}

\cvprfinalcopy 

\def\cvprPaperID{*} 

\begin{document}

\title{Rethinking Motion Representation: \\Residual Frames with 3D ConvNets for Better Action Recognition}

\author{Li Tao, Xueting Wang, Toshihiko Yamasaki\\
The University of Tokyo\\
{\tt\small {\{taoli, xt\_wang, yamasaki\}}@hal.t.u-tokyo.ac.jp}
}

\maketitle

\begin{abstract}
   Recently, 3D convolutional networks yield good performance in action recognition. However, optical flow stream is still needed to ensure better performance, the cost of which is very high. In this paper, we propose a fast but effective way to extract motion features from videos utilizing residual frames as the input data in 3D ConvNets. By replacing traditional stacked RGB frames with residual ones, 20.5\% and 12.5\% points improvements over top-1 accuracy can be achieved on the UCF101 and HMDB51 datasets when trained from scratch. Because residual frames contain little information of object appearance, we further use a 2D convolutional network to extract appearance features and combine them with the results from residual frames to form a two-path solution. In three benchmark datasets, our two-path solution achieved better or comparable performances than those using additional optical flow methods, especially outperformed the state-of-the-art models on Mini-kinetics dataset. Further analysis indicates that better motion features can be extracted using residual frames with 3D ConvNets, and our residual-frame-input path is a good supplement for existing RGB-frame-input models. 
\end{abstract}

\section{Introduction}

For action recognition, motion representation is an important challenge to extract motion features among multiple frames. Various methods have been designed to capture the movement.
2D ConvNet based methods use interactions in the temporal axis to include temporal information~\cite{karpathy2014large,wang2016temporal,li2019temporal,tsm,wang2018non}. 3D ConvNet based methods improved the recognition performance by extending 2D convolution kernel to 3D, and computations among temporal axis in each convolutional layers are \textit{believed} to handle the movements~\cite{c3d,p3d,s3d,i3d,res3d,r3d}.
State-of-the-art methods showed further improvements by increasing the number of used frames and the size of the input data as well as deeper backbone networks~\cite{slowfast,crasto2019mars,tran2019video}.

\begin{figure}[t]
    \begin{center}
    \includegraphics[width=0.9\linewidth]{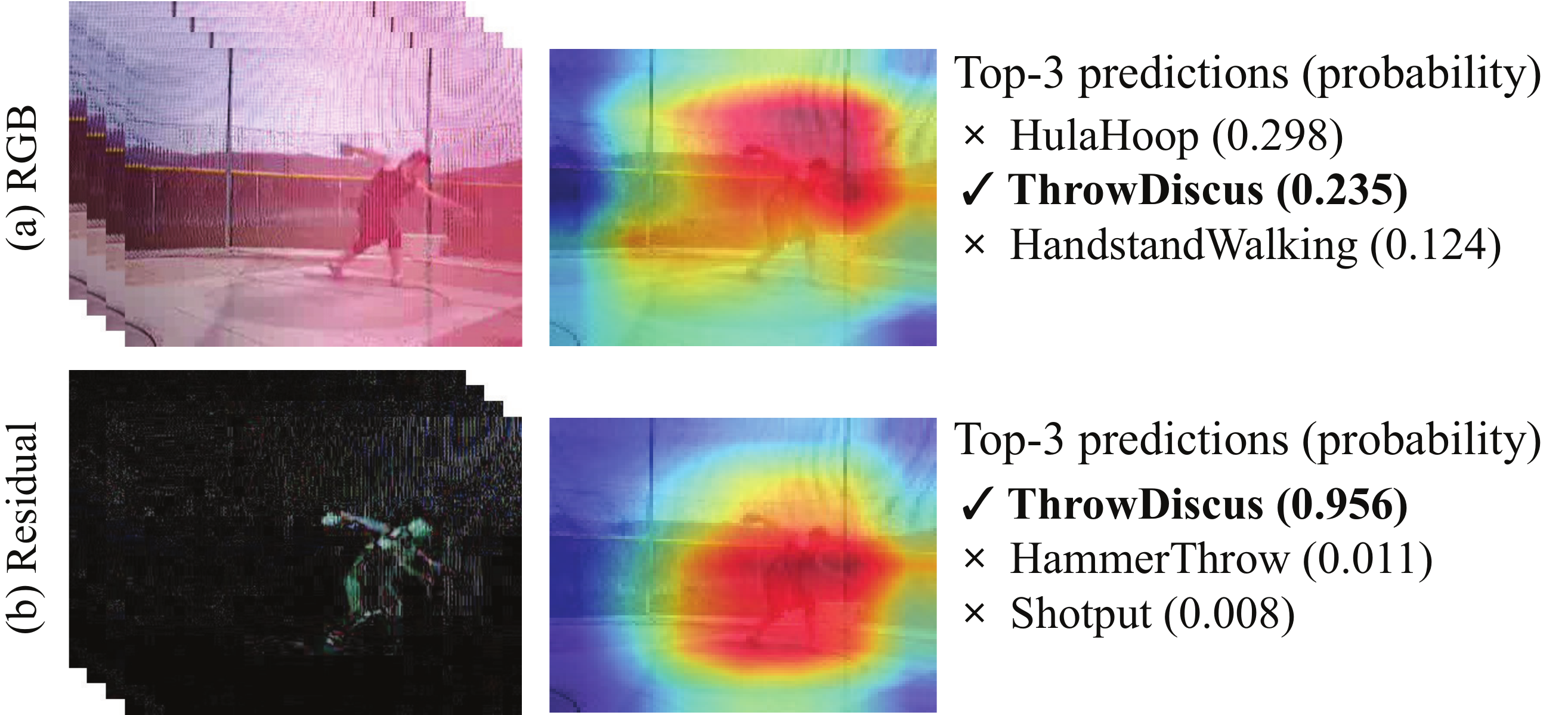}
    \end{center}
       \caption{An example of our residual frames compared with normal 3D ConvNet inputs. The residual-input model focused on the movement part while RGB-input model paid more attention on background, which lead to lower accuracy for prediction.}
    \label{fig:res_frame}
    \vspace{-5pt}
 \end{figure}

In a typical implementation of 3D ConvNets, these methods used stacked RGB frames as the input data. However, this kind of input is considered not enough for motion representation because the features captured from the stacked RGB frames may pay more attention to the appearance feature including background and objects rather than the movement itself, as shown in the top example in Fig.~\ref{fig:res_frame}. Thus, combining with an optical flow stream is necessary to further represent the movement and improve the performance, such as the two-stream models~\cite{feichtenhofer2016convolutional,feichtenhofer2016spatiotemporal,simonyan2014two}. However, the processing of optical flow greatly increases computation time\footnote{Because there are many types of implementation of optical flow, we do not refer to any specific type of implementation. But the calculation of optical flow is generally very expensive.}. Besides, obtaining two-stream results activation of the optical flow stream only after the optical flow data are extracted, which causes high latency.

In this paper, we propose an effective strategy based on 3D convolutional networks to pre-process RGB frames for the generation and replacement of input data. Our method retains what we call \textbf{residual frames}, which contain more motion-specific features by removing still objects and background information and leaving mainly the changes between frames. Through this, the movement can be extracted more clearly and recognition performance can be improved comparing to just using stacked RGB inputs as shown in the bottom sample in Fig.~\ref{fig:res_frame}. Our experiments reveal that our approach can yield significant improvements over \text{top-1} accuracies when those ConvNets are trained from scratch on UCF101~\cite{ucf101} and HMDB51~\cite{hmdb} datasets. One may think that our approach is naive and therefore cannot be applied to videos with global motion, but this will also be addressed in Section~\ref{global_motion}.

For larger action recognition datasets such as Mini-kinetics~\cite{s3d} and Kinetics~\cite{kinetics}, the definitions of the actions become more complex such as \textit{Yoga} containing various combination of simple actions, and these datasets have a large amount of compound labels, such as \textit{playing guitar} and \textit{playing ukulele}, where the movement is almost the same and the difference is mainly on the objects.
In this case, it is difficult to distinguish by only motion representation without enough appearance features. Therefore, we propose a two-path solution, which combines the residual input path with a simple 2D ConvNet to extract appearance features from a single frame. Experiments show that our proposed two-path method obtains better performance over some two-stream models on UCF101 / HMDB51 / Mini-kinetics datasets when using the same input shapes and similar or even shallower network architectures. 

Our contributions are summarized as follows:
\begin{itemize}
  \setlength{\parskip}{-3pt}
\item We propose a simple, fast, but effective way for 3D ConvNets to better extract motion features by using stacked residual frames as the model input.
\item The proposed two-path solution including a 3D ConvNet with residual input as the motion path and a 2D ConvNet as the appearance path can achieve better performance than other methods using similar settings.
\item  Our proposal can avoid the requirement of high-cost computation for optical flow while ensuring high performance. Our analysis also suggests potential limitations in the current action recognition task.
\end{itemize}

We would like to clarify that we are proposing a new way for motion representation. For this purpose, we do not always focus on the better performance than other approaches based on very deep and complex DNN architectures as well as other training / parameter settings. Instead, we discuss why and how much our approach is reasonable as compared to optical-flow-based and RGB-only approaches. We will release our code if the paper is accepted.

\section{Related works}
In this section, traditional action recognition networks are introduced. Though temporal modeling usually exists among those networks, we use another subsection to introduce and discuss this in detail because temporal information is a key feature. Model combination is set as another subsection to clearly see the solution route maps for high accuracies.

\subsection{Deep action recognition}
\noindent{\bf 2D solution.} 2D ConvNets based methods mainly consist of frame-level feature representation and temporal modeling to fuse these features. When treating each frame of a video as a single image, 2D ConvNets which are effective for image classification task can be directly applied to video recognition. Karpathy \etal~\cite{karpathy2014large} tried different ways to fuse features from 2D ConvNet and then used fused features to classify videos. Temporal Segment Networks (TSN)~\cite{wang2016temporal} was designed to extract average features from stride sampled frames. Two-stream ConvNets~\cite{simonyan2014two, feichtenhofer2016convolutional, feichtenhofer2016spatiotemporal} used an additional optical flow stream. And for both RGB stream and optical flow stream, 2D ConvNets were used. Recent works such as Temporal Bilinear Networks (TBN)~\cite{li2019temporal} and Temporal Shift Module (TSM)~\cite{tsm} are variants of 2D ConvNets. Compared to 3D counterparts, 2D methods are more efficient because fewer parameters are used, and the performance is highly related to the temporal modeling. Our method uses a 2D network to extract appearance features considering the high efficiency of 2D models, and the proposed appearance path uses less input than existing 2D ConvNets, which is more efficient.\\

\noindent{\bf 3D solution.} 3D ConvNets based methods directly use 3D convolution kernels to process input video frames. The computation between frames is carried out when the temporal kernel size is 2 or larger, and spatial-temporal features can be automatically learned by network optimization. Tran \etal~\cite{c3d} proposed C3D, which consists of 8 directly-connected convolutional layers and 2 fully-connected layers. Hara \etal~\cite{res3d} conducted many experiments on the 3D version of residual networks, including different depths and using some variants such as ResNeXt~\cite{resnext}. Carreira \etal~\cite{i3d} proposed I3D based on Inception network. SlowFast~\cite{slowfast} used two ResNet pathways to capture multi-scale information in the temporal axis. Despite of different network architectures, 3D convolutional kernel also has variants. One $k \times k \times k$ kernel can be separated into two parts, $k \times 1 \times 1$ and $1 \times k \times k$. Based on this, P3D~\cite{p3d}, R(2+1)D~\cite{r3d}, and S3D~\cite{s3d} were proposed. The backbones of mainstream networks are ResNets~\cite{resnet} and Inception network~\cite{inception}. Neural architecture searching (NAS) is used in~\cite{nas} to get efficient network architectures. However, because the parameter number is larger than 2D counterparts, 3D models are prone to overfitting when trained from scratch on small datasets such as UCF101~\cite{ucf101} and HMDB51~\cite{hmdb}. Fine-tuning models pre-trained on very large dataset such as Kinetics~\cite{kinetics} is one solution to acquire good performance on these small datasets. From another point of sight, our proposed method focus more on the movement itself and utilize  a 3D ConvNet with higher motion representation ability by using residual frames as input. In this way, we can reduce the tendency to over-fit on small datasets compared to normal RGB inputs when using the same network architectures. 

\subsection{Temporal modeling}
For 2D ConvNets, some models~\cite{karpathy2014large,wang2016temporal} have been proposed which simply averaged frame features to represent videos. Donahue \etal~\cite{lstm} used 2D models to extract features using long short-term memory (LSTM)~\cite{lstm_video}. Zhou \etal~\cite{zhou2018temporal} proposed Temporal Relation Network to learn temporal dependencies. Temporal Bilinear Networks~\cite{li2019temporal} uses temporal bilinear modeling to embed temporal information. Temporal Shift Module~\cite{tsm} shifts 2D feature maps along temporal dimension.

For 3D ConvNets, temporal modeling is automatically processed by learning kernels in the temporal axis. Because 3D ConvNets use stacked RGB frames as input, the computation among frames is believed to learn motion features, while the spatial computation is for spatial feature embedding. Therefore, existing 3D models do not pay much attention to this part, and trusting the capabilities of network. Recently, Crasto \etal~\cite{crasto2019mars} trained a student network using RGB-frame input by learning feature representation from a teacher network, which had been trained using optical flow data to enhance temporal modeling.  

Our proposed two-path method consists of an appearance path using a 2D ConvNet only to extract appearance features and a motion path using 3D ConvNet to calculate motion features. Temporal modeling only exists in the motion path. The use of residual frames differs from using as motions exist not only in the temporal dimension of residual frames, but also in the spatial dimension because one residual frame is generated from two adjacent frames. 

\subsection{Two-stream model}
Two-stream models usually stand for those methods combining 2D features / results from RGB stream with optical flow stream~\cite{simonyan2014two, feichtenhofer2016convolutional, feichtenhofer2016spatiotemporal}. Some researchers extended the concept by combining RGB-frame-input path with another path which uses pre-computed extra motion features, such as trajectories~\cite{idt} or SIFT-3D~\cite{sift3d}, as well as optical flow. Many existing methods can then be extended by combining motion feature stream to further improve their performances~\cite{crasto2019mars,i3d,r3d}. To distinguish our proposal from the aforementioned two-stream methods, we refer to our method as `\textbf{two-path}' rather than `two-stream' because we do not use any pre-computed motion features.

\section{Proposed method}

In this section, we first introduce our proposal that uses residual frames as a new form of input data for 3D ConvNets. Because residual frames lack enough information for objects, which are necessary for the compound phrases used for label definitions in most video recognition datasets, we further propose a two-path solution to utilize appearance features as an effective complement for motion features learned from the residual inputs.

\subsection{Residual frames}
For 3D ConvNets, stacked frames are set as input, and the input shape for one batch data is $T \times H \times W \times C$, where $T$ frames are stacked together with height $H$ and width $W$, and the channel number $C$ is 3 for RGB images. We denote the data as $THW$ for simplicity. The convolutional kernel for each 3D convolutional layer is also in 3 dimensions, being $k_T \times k_H \times k_W$. Then for each 3D convolutional layer, data will be computed among three dimensions simultaneously. However, this is based on a strong assumption that motion features and spatial features can be learned perfectly at the same time. To improve performance, many existing 3D models expand weights from 2D ConvNets to initialize 3D ConvNets, and this has been proved to provide higher accuracies. Pre-training on larger datasets will also enhance performance when fine-tuned on small datasets.

When subtracting adjacent frames to get a residual frame, only the frame differences are kept. In a single residual frame, movements exist in the spatial axis. Using residual frames for 2D ConvNets have been attempted and proved to be somewhat effective~\cite{wu2018compressed}. However, because actions or activities are complex with much longer durations, stacked frames are still necessary. In stacked residual frames, the movement does not only exist in the spatial axis, but also in the temporal axis, which is more suitable for 3D ConvNets because 3D convolution kernels will process data in both spatial and temporal axes. Using stacked residual frames helps 3D convolutional kernel to concentrate on capturing motion features because the network does not need to consider the appearance information of objects or backgrounds in videos. 

Here we use $frame_i$ to represent the $i_{th}$ frame data, and $Frame_{i\sim j}$ denotes the stacked frames from the $i_{th}$ frame to the $j_{th}$ frame. The process to get residual frames can be formulated as follows,

\vspace{-15pt}
\begin{equation}
   ResFrame_{i\sim j} = | Frame_{i\sim j} - Frame_{i+1\sim j+1} |
   \vspace{-5pt}
\end{equation}
The computational cost is cheap and can even be ignored when compared with the network itself or optical flow calculation.

With this change, 3D ConvNet can extract motion features by focusing on the movements in videos alone. However, by ignoring objects and backgrounds, some movements in similar actions become indistinguishable. For example, in the actions \textit{Apply Eye Makeup} and \textit{Apply Lipstick}, the main difference lies in the location of the movement being around the eyes or the mouth rather than the movement itself. In this example, 3D ConvNets may be able to distinguish them to some extent but the loss of information does increase the difficulty. Therefore, we use a 2D ConvNet to process the lost appearance information and combine with a 3D ConvNet using residual frames as input to form a two-path network.

\subsection{Two-path network}

\begin{figure}[t]
    \begin{center}
    \includegraphics[width=0.9\linewidth]{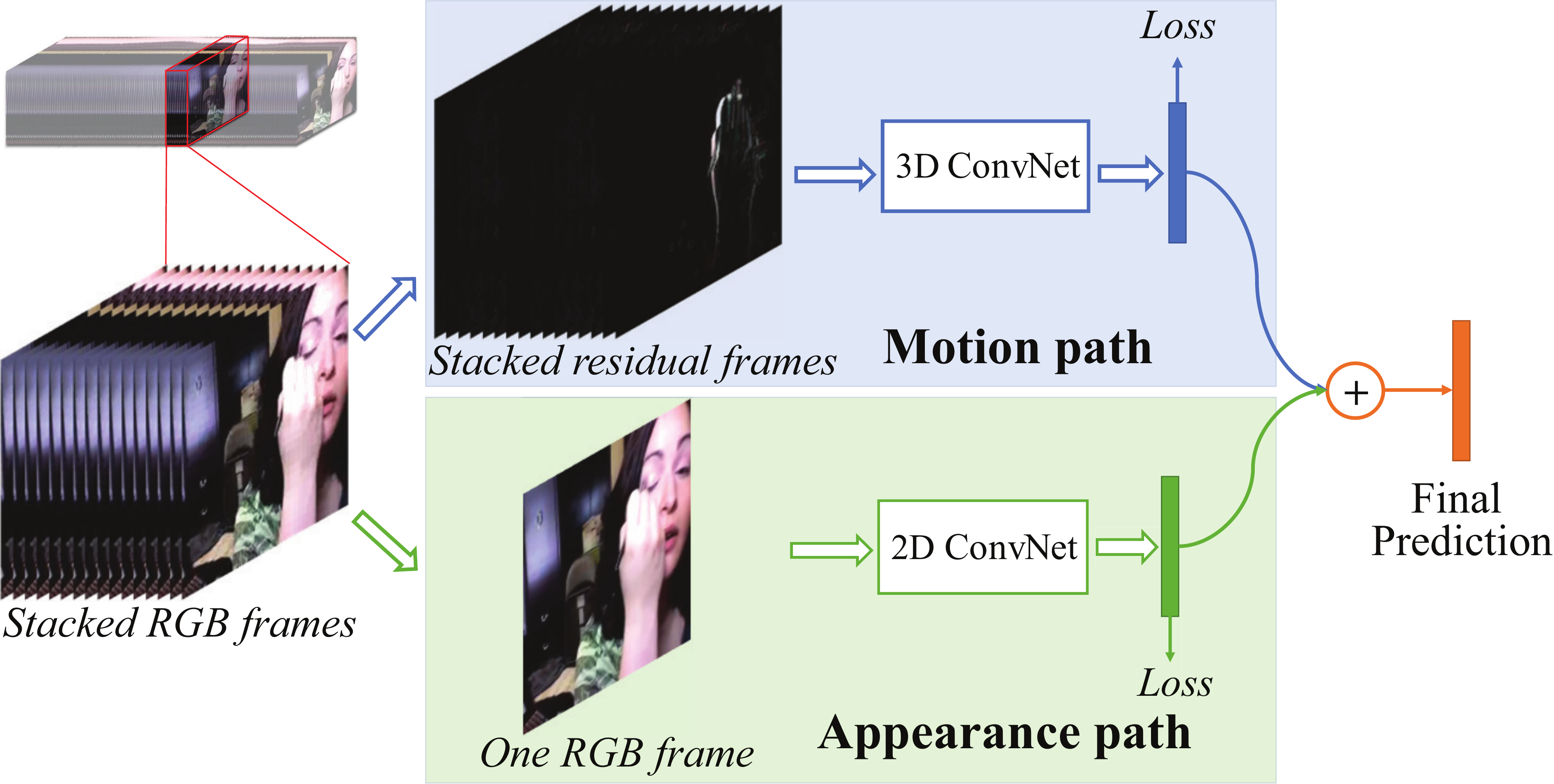}
    \end{center}
    \vspace{-5pt}
       \caption{Framework of our two-path network. The motion path and the appearance path are trained separately using cross-entropy loss. Action recognition is carried out within each path. In inference period, the output probabilities from two paths are averaged. In this way, both motion features and appearance features are utilized for final classification.}
    \label{fig:twopath}
    \vspace{-10pt}
 \end{figure}

Our two-path network is formed by a motion path and an appearance path, which is illustrated in Fig.~\ref{fig:twopath}

\noindent{\bf Motion path.} 
Because residual frames are used in this path, movements then exist in both spatial axis and the temporal axis. Therefore, 3D convolution layers are used in this path. Because there are many existing 3D convolution based network architectures which have been proved effective in many action recognition datasets, we do not focus on designing a new network architecture in this paper. To verify the robustness and versatility of our proposal, we conduct experiments on various models, and discussed especially on ResNet-18-3D as its good performance. In the original ResNet-18-3D~\cite{res3d}, convolution with stride is used at the beginning of several residual blocks to perform down-sampling. We attempt another version of residual blocks which uses max-pooling layers at the end of each corresponding blocks. These two versions have almost the same network parameters. 

\noindent{\bf Appearance path.} 
By using residual frames with 3D ConvNets, motion features can be better extracted, while background features which contains object appearances are lost. The lost part can be extracted by a 2D ConvNet, which uses one RGB frame as input. The goal for our appearance path is to embed object and background appearances which are mostly lost in the motion path. Therefore, in contrast to TSN or other complex models, a simple 2D ConvNet is sufficient. The naive 2D ConvNet treats action recognition as a simple image classification problem. During training, only one frame in a video is randomly selected in one epoch. 

For the combination of these two paths, we average the predictions for the same video sample. There are early fusion methods that may be more effective, which we leave as our future work.

\section{Experiments}
\subsection{Datasets and metrics}
\noindent{\bf Datasets.} 
There are several commonly used datasets for video recognition tasks. Thanks to the large amount of videos and labels in these datasets, deep learning methods can detect a high number of actions. We mainly focus on the following benchmarks: UCF101~\cite{ucf101}, HMDB51~\cite{hmdb}, and Kinetics400~\cite{kinetics}. UCF101 consists of 13,320 videos in 101 action categories. HMDB51 is comprised of 7,000 videos with a total of 51 action classes. Kinetics400 is much larger, consisting 400 action classes and contains around 240k videos for training, 20k videos for validation and 40k videos for testing. For the Kinetics400 dataset, because it is very large, we mainly perform our experiments on its subset, Mini-kinetics~\cite{s3d}, which consists of 200 action classes with 80,000 videos for training and 5,000 videos for validation. The actual data used in our experiments may be a little smaller because some videos were unavailable.

\noindent{\bf Metrics.} 
We report all experiments with top-1 and top-5 accuracies for all experiments. The performance on Mini-kinetics was evaluated on the validation split. We also use correlation coefficient index for deeper analysis between different models, which may indicate the relationships between the knowledge learned from existing models.

\subsection{Scratch training and fine-tuning}
There are always two ways to train a network, either training from scratch or fine-tuning from a pre-trained one. There is an obvious gap between these two training routes. Thanks to the proposal of the Kinetics datasets, several 3D convolution based models have been proposed with better performances using pre-trained models. Therefore, many recent works based their results on fine-tuned models for small datasets such as HMDB51 and UCF101, and trained from scratch for larger datasets such as Kinetics400 and its subset, Mini-kinetics.

Models can benefit from larger datasets, but training on larger datasets significantly increases computation time, so repeatedly increasing the size of datasets to improve performance is not always a solution. In this paper, in addition to the default settings discussed above, we also look into the situation that no additional knowledge is available. Specifically, we want to explore the limitations for 3D ConvNets on UCF101 and HMDB51 without any additional datasets. 

\subsection{Implementation details}
\noindent{\bf Motion path.} In this path, stacked residual frames are set as the network input data. Residual frames are used identically to traditional RGB frame clips. For 3D ConvNets in action recognition, there are several input setting choices. 3D ConvNets started from~\cite{c3d} which used a clip of $16$ consecutive frames, with a $112 \times 112$ slice cropped in the spatial axis. To achieve the state-of-the-art results, clips in size $64\times 224\times 224$ were used in many recent works. When using such a large input data size, improvements can be achieved but limited while longer training time as well as larger memory occupations are necessary. Therefore, for all of our motion path, following~\cite{c3d}, frames are resized to $170\times 128$ and $16$ consecutive frames are stacked to form one clip. Then, random spatial cropping is conducted to generate an input data of size $16\times 112\times 112$. Before it is fed into the network, random horizontal flipping is performed. Jittering along the temporal axis is applied during training. The backbone in most our experiments is ResNet-18-3D. We tried two variants of ResNet-18-3D, the difference of which is whether using convolution with strides at the beginning of some residual blocks or using max-pooling at the end of corresponding blocks instead. R(2+1)D, I3D, and S3D are also tested to verify the robustness of our proposal. The batch size is set to 32. When models are trained from scratch, the initial learning rate is set to 0.1. We trained models for 100 epochs on UCF101 and HMDB51, and used 200 epochs for Mini-kinetics. When fine-tuning on UCF101 and HMDB51 using Kinetics400 pre-trained models, model weights are from~\cite{res3d} and the network architecture remains the same as~\cite{res3d}. The initial learning rate became 0.001, and 50 epochs were sufficient.

\noindent{\bf Appearance path.} In contrast to TSN, our appearance path used a simpler model which treats action recognition as image classification because appearances in consecutive frames changed infrequently, and the goal for this path is to capture appearance features for background and objects. Frames are first resized to $256\times 256$ and random spatial cropping and random horizontal flipping are applied in sequence to generate input data with a size of $224\times 224$. This progress is standard in image classification to enable the use of many pre-trained models. ResNet-18, ResNet-34, ResNet-50, and ResNeXt-101 were used to test the impact of different model depth. In addition, models were also trained from scratch to see the performances when no additional knowledge is provided.

\noindent{\bf Training recipes.} 
We noticed that few works paid attention to training recipes in video recognition. We used several training recipes to train our models. Specifically, we tried to use different activation functions and different learning rate decay methods. Experiments for ths part are mainly carried out on the UCF101 dataset. For larger datasets, we find that some settings still work, and we think they can be called as `bag of tricks' in video recognition tasks. 

\noindent{\bf Testing and Results Accumulation.} 
There are two means of testing for action recognition using 3D ConvNets. One is to uniformly get video clips from one video, which means a fixed number of clips is generated and set as the input of the model, regardless of the video length. The predictions are averaged over all video clips to generate the final result. The other method uses non-overlapping video clips, which means longer videos will produce more video clips. The final result for one video is also generated by averaging these video clips. We performed a small test for these two means of testing and found the difference can be ignored because all of the clip results are averaged in both methods. Thus, we used the uniform method in our experiments, and our appearance path used a fixed number frames sampled from all video frames to match the motion path.

\section{Results and discussion}
In this section, results from single paths are first introduced. The motion path is used to investigate the effectiveness of stacked residual frames. Then, results from the appearance path are reported. Further analysis is conducted to explore the connections between models, especially RGB 2D model and RGB / residual 3D model. Finally, we show the performance of our proposed two-path network comparing to various existing models.

\subsection{Single path}\label{global_motion}
\noindent\textbf{Motion path.}
For the motion path, different training recipes were investigated first. Different activation functions were tried and we found that, in contrast to existing 3D convolution based methods~\cite{res3d,r3d,s3d,c3d,p3d,i3d} which use ReLU as the default activation function, replacing ReLU with ELU improved the \text{top-1} accuracy by 2.6\% points (from 51.9\% to 54.5\%) and 3.3\% points (from 58.0\% to 61.3\%) for two experimental versions (\textit{convolution with stride} and \textit{max-pooling}) of ResNet-18 (Table~\ref{table:result_ucf3d}). Similar results can be found for Mini-kinetics. To get better performance, we use ResNet-18 with max-pooling layers as our default model version.

Compared to RGB clips, stacked residual frames maintain movements in both spatial and temporal axes, which takes greater advantage of 3D convolution. Results are shown in Table~\ref{table:result_ucf3d} and the following discussion is all based on this table. By simply replacing RGB clips with our proposed residual clips, ResNet-18-3D results can be improved from 51.9\% to 72.4\%. To the best of our knowledge, this outperforms the current state-of-the-art results when models are trained from \textbf{scratch} on UCF101. In addition to directly using residual frames, feature differences were carried out. In Model \textit{ResNet-18(fea\_diff)}, we used RGB clips as input while calculating feature differences in the temporal axis after first convolutional layer. The results were then fed into the rest of the network. However, this produced lower accuracies than directly using the residual frames as the network input. R(2+1)D, I3D, and S3D are also experimented and improvements are achieved by more than 10\% when replacing original RGB input with our residual frames. 

To sum up our residual inputs, we can see that this approach is robust for different model architectures. Because ResNet-18 is light-weighted and has good performance, we used ResNet-18 as the default backbone in our motion path.

\begin{table}[t]
    \begin{center}
    \scalebox{0.7}{
   \setlength{\tabcolsep}{1.8mm}{
   \begin{tabular}{|c|c|c|c|c|c|c|}
        \hline
        Model & type & recipe & \textbf{residual} & top-1 & top-5 \\
        \hline\hline
        ResNet-18 baseline~\cite{res3d} & - & - & - & 42.4 & - \\
        STC-ResNet-101~\cite{stc} & - & - & - & 47.9 & - \\
        NAS~\cite{nas} & - & - & - & 58.6 & -\\
        \hline
        ResNet-18 & stride & $\times$ & $\times$ & 51.9 & 76.3 \\
        ResNet-18 & pool & $\times$ & $\times$  & 58.0 & 82.6 \\
        ResNet-18 & stride & \checkmark & $\times$   & 54.5 & 80.6 \\
        ResNet-18 & pool & \checkmark & $\times$  & 61.3 & 84.2 \\
        ResNet-18 (fea\_diff) & pool & \checkmark & \checkmark  & 64.7 & 86.6 \\
        ResNet-18 & stride & \checkmark& \checkmark  & 66.4 & 88.0 \\
        \textbf{ResNet-18} & pool & \checkmark & \checkmark & \textbf{72.4} & \textbf{89.7} \\
        \hline
        R(2+1)D~\cite{r3d} & stride & $\times$ & $\times$ & 51.8 & 79.2 \\
        R(2+1)D~\cite{r3d} & stride & \checkmark & \checkmark & \textbf{66.7} & \textbf{88.3} \\
        \hline
        I3D~\cite{i3d} & - & $\times$ & $\times$ & 56.5 & 81.3 \\
        I3D~\cite{i3d} & - & \checkmark & \checkmark & \textbf{66.6} & \textbf{87.0} \\
        \hline
        S3D~\cite{s3d} & - & $\times$ & $\times$ & 51.1 & 77.4 \\
        S3D~\cite{s3d} & - & \checkmark & \checkmark & \textbf{64.8} & \textbf{86.9} \\
        \hline
    \end{tabular}}}
    \end{center}
    \caption{Different settings on UCF101 \textit{split}~1, all models are trained from \textbf{scratch}. The original ResNet-18 baseline did not search better training recipes. Our implement results are higher than the baseline using the same network architecture. We also implement R(2+1)D, I3D and S3D and more than 10\% improvement can be achieved for each by using our residual input.}
	\label{table:result_ucf3d}
\end{table}

\begin{table}[tb]
    \begin{center}
    \scalebox{0.7}{
    \setlength{\tabcolsep}{1.1mm}{
   \begin{tabular}{|c|c|c|c|c|c|}
        \hline
        Model & Type & Pre-train & UCF101 & HMDB51 & Mini-kinetics \\
        \hline\hline
        ResNet-18 & RGB & $\times$ & 51.9 & 22.2 & \textbf{65.0} \\
        ResNet-18 & Residual & $\times$ & \textbf{72.4}& \textbf{34.7} & 64.4 \\
        \hline
        ResNet-18 & RGB & \checkmark & 84.4 & \textbf{56.4} & - \\
        ResNet-18 & Residual & \checkmark & \textbf{89.0} & 54.7 & - \\
        \hline
    \end{tabular}}}
    \end{center}
    \caption{Top-1 results for motion path on three benchmark datasets. Training on Kinetics400 costs too much time. Therefore, for fine-tuning models, we used pre-trained models provided in~\cite{res3d}, the only difference is that we use our residual input. The reported results are on UCF101 \textit{split}~1 and HMDB51 \textit{split}~1.}
	\label{table:hmdb_k400}
\end{table}

\begin{figure}[t]
    \begin{center}
    \includegraphics[width=0.85\linewidth]{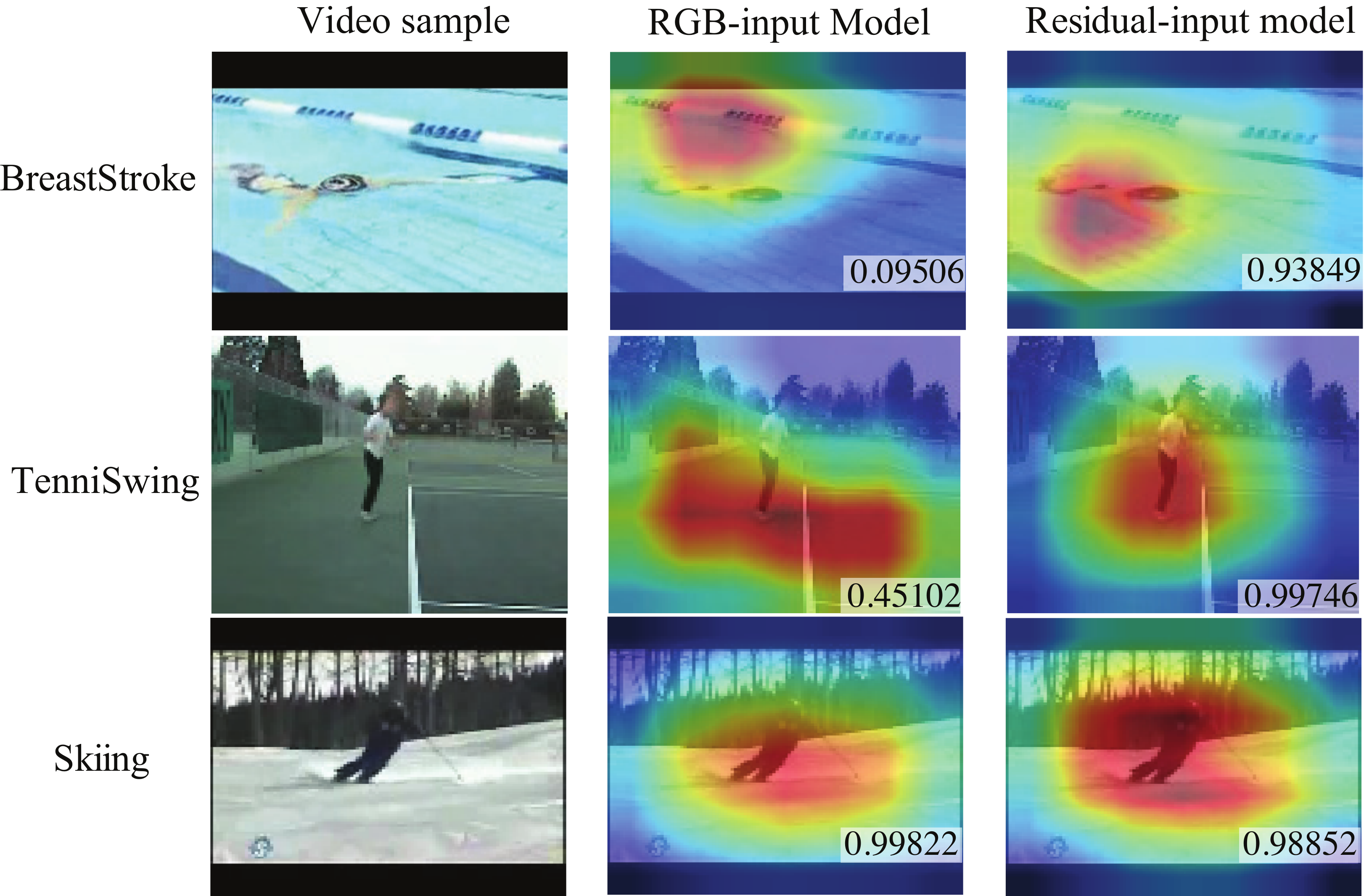}
    \end{center}
       \caption{Visualization using grad-cam~\cite{gradcam}. The number is the corresponding prediction probability for each sample. Residual-input model focused more on the moving entity and the moving area while RGB-input included more background information.}
    \label{fig:gradcam}
    \vspace{-10pt}
 \end{figure}

We also tested the performance on HMDB51 and Mini-kinetics. Results are shown in Table~\ref{table:hmdb_k400}. On HMDB51 \textit{split}~1, the results can be improved from 22.2\% to 34.7\% when replacing the original input with residual frames. However, the improvement can not be observed for Mini-kinetics because the labels are more related to objects rather than actions, which is the main reason of introducing our appearance path. Residual-input model can also benefit from pre-trained models when fine-tuning, yielding 89.0\% on UCF101 \textit{split}~1. The results on HMDB51 are not as good as RGB model because on this dataset, the range of one variation of one action is larger. For example, the category \textit{Dive} including bungee jumping and a movement by a score keeper on the ground. And many movements are inconsistent in one category while the samples are few, which greatly increases the difficulty for residual inputs.

\begin{figure*}[t]
    \begin{center}
    \includegraphics[width=0.9\linewidth]{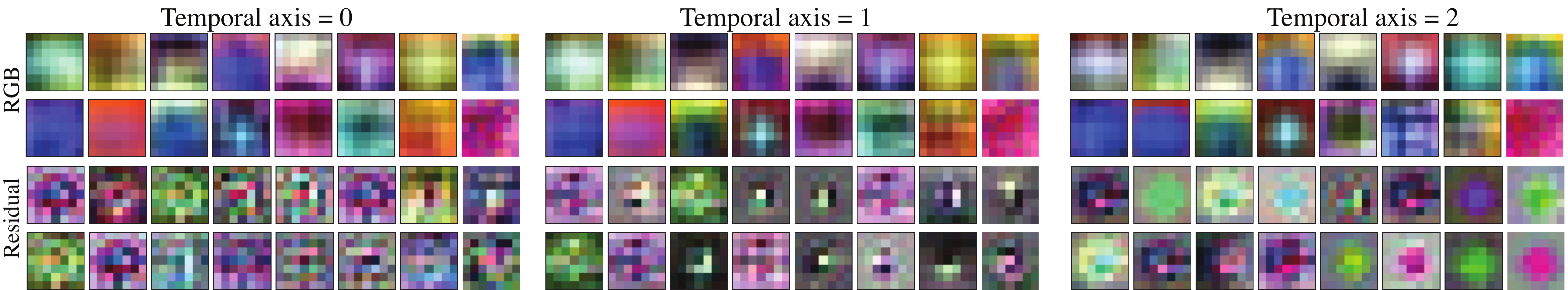}
    \end{center}
       \caption{Visualization for model weights. Models are trained from \textbf{scratch} on Mini-kinetics. Filters in RGB-input model are similar among temporal axis. On the other hand, in the residual-input model, the weights indicate that the residual-input model will be more sensitive for changes in temporal dimension.}
    \label{fig:filters}
    \vspace{-10pt}
 \end{figure*}

 \begin{figure}[t]
    \begin{center}
    \includegraphics[width=0.85\linewidth]{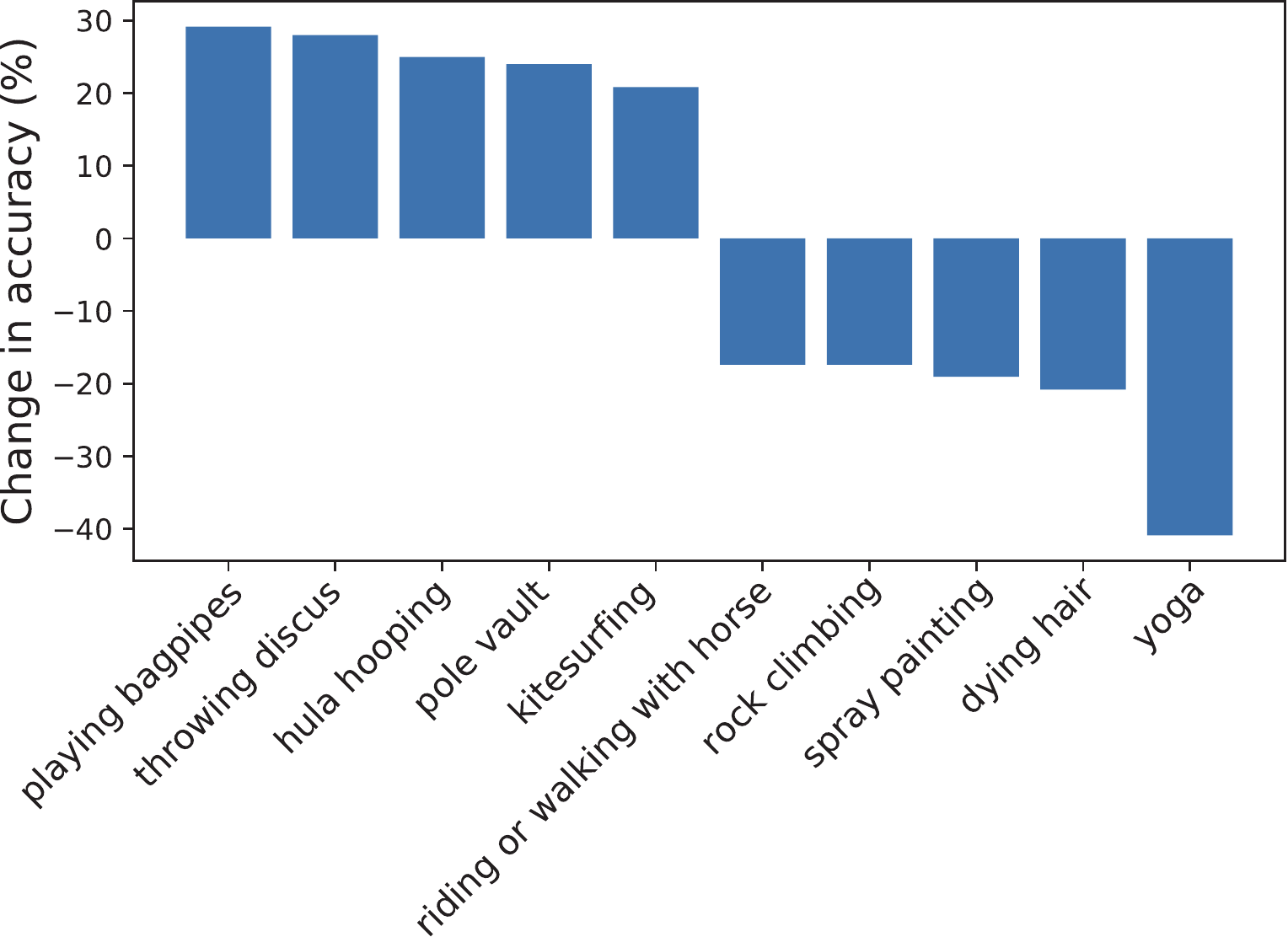}
    \end{center}
    \vspace{-5pt}
    \caption{Accuracy difference between models with residual inputs and RGB inputs on Mini-kinetics. Best-5 and worst-5 categories are illustrated.}
    \label{fig:difference}
 \end{figure}

For deeper analysis, we further use grad-cam~\cite{gradcam} for visualization. As shown in Fig.~\ref{fig:gradcam}, the residual-input model pays attention to the action entity while the RGB-input model focuses more on the background. The prediction probability is low for \textit{BreastStroke} because RGB model gives higher probability for another swimming style \textit{FrontCrawl}. 

The first 16 out of 64 convolutional filters in $conv1$ layers from the RGB-input model and the residual-input model are illustrated in Fig.~\ref{fig:filters}. These two models are both trained from scratch on Mini-kinetics. We can see that the filters in the RGB-input model are similar among different temporal axes. For this reason, using ImageNet~\cite{imagenet} pre-trained models can achieve good performance, even when using naive 2D models, while with our residual inputs, the performance is a little bit lower. The filters in residual input model differs from each other among different temporal axis, indicating that this model is more sensitive for changes in time. The accuracy differences between our residual-input model and the RGB-input model are illustrated in Fig.~\ref{fig:difference}. We show the best-5 and worst-5 classes. The positive peak belongs to the class \textit{playing bagpipes} and we found that in this category, there are global movements cause by lens shake and other irrelevant movements by bystanders while our residual-input model can handle this kind of movement. Movements in \textit{throwing discus} and \textit{hula hooping} are highly consistent. In contrast, movements in \textit{yoga} varies from each other while the appearance information plays a more important role. 

Based on our analysis, the ability of 3D ConvNets may be limited because of the ambiguity in action labels. Additionally, more attention is paid to appearance rather than movements for RGB 3D models. 

\noindent\textbf{Appearance path.} 
For the appearance path, four ResNet architectures were used, namely ResNet-18, ResNet-34, ResNet-50, and ResNeXt-101. Scratch training as well as fine-tuning from ImageNet pre-trained models were both tried. The results are shown in Table~\ref{table:result_2d}.

\begin{table}[t]
    \begin{center}
   \scalebox{0.7}{
   \setlength{\tabcolsep}{0.8mm}{
   \begin{tabular}{|c|c|c|c|c|c|c|}
        \hline
        Dataset & \multicolumn{2}{|c|}{UCF101} & \multicolumn{2}{|c|}{HMDB51} & \multicolumn{2}{|c|}{Mini-kinetics}\\
        \hline\
        Pre-train  & Scratch & ImNet & Scratch & ImNet & Scratch &ImNet \\
        \hline\hline
        ResNet-18 & 37.7 & 79.6 & 25.0 & 42.6 & 57.7 &64.4\\
        ResNet-34 & 40.1 & 81.5 & 24.8 & 43.1 & 59.4 & 68.9\\
        ResNet-50 & 33.7 & 83.8 & 21.3 & 43.4 & 58.6 & 69.7\\
        ResNeXt-101 & 34.4 & 85.2 & 23.3 & 45.6 & 59.7 & 70.5\\
        \hline
    \end{tabular}}}
    \end{center}
    \caption{Top 1 accuracies using appearance path on UCF101 \textit{split}~1, HMDB51 \textit{split}~1, and Mini-kinetics. Models are trained either from scratch or by using fine-tuning.}
    \label{table:result_2d}
    \vspace{-10pt}
\end{table}

We can clearly find that the gap is large for 2D ConvNets between these two training ways, which is consistent with previous works on image classification tasks. However, pre-training also needs much time if no pre-trained models are provided. For better performance, deeper networks generally provide higher scores. 

Regarding Mini-kinetics, ImageNet pre-trained models were directly used and high accuracies could be achieved. Among these 2D ConvNets, the best \text{top-1} accuracy was 70.5\% points which is very high. However, in this case, the action recognition task is treated as a simple image classification task, which does not benefit from the use of any temporal information. 

The performance of ResNet-18-2D using pre-trained weights is 79.6\%, which is close to the performance of scratch training using ResNet-18-3D in Table~\ref{table:result_ucf3d}, 72.4\%. Though it may be unfair to compare these two models because the 2D version utilizes image classification knowledge to initialize its parameters while the 3D version does not. Duplicating ImageNet pre-trained model parameters in 3D ConvNets could be a good solution. However, it is still prone to mainly using appearance features.

\noindent\textbf{Analysis among models.}
The difference between 2D convolution and 3D convolution is that 3D convolution has another dimension which is aimed to process temporal information. For continuous frames, especially those trimmed videos provided in video recognition datasets, the difference between frames is limited. Therefore, the 3D convolution may not process temporal information efficiently. Duplicating ImageNet per-trained model parameters as the initial model parameters does provide improvements, but then spatial-temporal convolution might be lazy during fine-tuning progress because even for models trained from scratch, model weights are tend to be similar among temporal axis~(Fig.~\ref{fig:filters}).

Here we introduce the correlation coefficient index to calculate the relationships between different models. 2D models and 3D models were tested. For 2D ConvNets, we also used optical flow stream as a comparative model. Correlation coefficient indexes for per-category accuracies between two different models will be reported in Table~\ref{table:corr}. The backbone networks are ResNeXt-101-2D and ResNet-18-3D. All models were fine-tuned to ensure the classification performance. From the table, we can see that the correlation coefficient index for RGB 2D and 3D models is high, which indicates that these two approaches may make judgement in a similar way while optical flow stream differs significantly. Our residual-input model has a high correlation with RGB 3D models because of the same network architecture. However, the correlation becomes lower with RGB 2D models because using residual frames results in more motions being used for classification rather than appearance.

\begin{table}[tb]
    \begin{center}
    \scalebox{0.7}{
    \setlength{\tabcolsep}{1.8mm}{
   \begin{tabular}{|c|c|c|c|c|}
        \hline
        \multicolumn{2}{|c|}{Model} & \multicolumn{2}{|c|}{Model} & \multirow{2}{*}{Correlation} \\
        \cline{1-4}
        Input  & Type & Input & Type &   \\
        \hline\hline
        RGB & 2D & RGB & 3D & \textbf{0.839} \\
        RGB & 2D & Residual & 3D & \textbf{0.663} \\
        RGB & 2D & Flow & 2D & 0.505 \\
        Flow & 2D & RGB & 3D & 0.569 \\
        Flow & 2D & Residual & 3D & 0.534 \\
        RGB & 3D & Residual & 3D & 0.791 \\
        \hline
    \end{tabular}}}
    \end{center}
    \caption{Correlation coefficient indexes on UCF101 \textit{split}~1. \textit{Type} means the type of convolution kernels used in the network.}
	\label{table:corr}
\end{table}

\subsection{Two-path network}
By combining the motion path with the appearance path, appearances and motions can be used to get the predictions. Because we have several models, we tried different combinations among different models. For example, in the UCF101 dataset, we try different combinations by selecting two models among the 2D ConvNet RGB model, the 2D ConvNet optical flow model, the 3D ConvNet RGB model and the 3D ConvNet residual model. The results are listed in Table~\ref{table:result_2path_ucf}. In our implementation, the optical flow path used a ResNeXt-101 backbone, which is the same as our appearance path. However, the combination of optical flow and other RGB models produces side effects on the accuracies. The identical values in this table are the result of rounding because the accuracies happen to be close enough to exceed the points of precision used in this table. 

\begin{table}[tb]
    \begin{center}
    \scalebox{0.7}{
    \setlength{\tabcolsep}{1.8mm}{
   \begin{tabular}{|c|c|c|c|c|c|}
        \hline
        \multicolumn{2}{|c|}{Model} & \multicolumn{2}{|c|}{Model} & \multirow{2}{*}{top-1} & \multirow{2}{*}{top-5} \\
        \cline{1-4}
        Input & Type & Input & Type & & \\
        \hline\hline
        RGB & 3D & Optical Flow & 2D & 75.7 & 92.1 \\
        RGB & 3D & Residual & 3D & 87.4 & 97.5 \\
        Residual & 3D & Optical flow & 2D & 75.7 & 92.1 \\
        RGB & 2D & Optical Flow & 2D & 75.7 & 92.1 \\
        RGB & 2D & RGB & 3D & 86.6 & 97.1 \\   
        RGB & 2D & Residual & 3D &  \textbf{90.3} & \textbf{98.5}\\
        \hline
    \end{tabular}}}
    \end{center}
    \caption{Results from different combination of different models on UCF101 \textit{split}~1. Our combination yielded the best performances.}
	\label{table:result_2path_ucf}
\end{table}

\begin{table}[tb]
    \begin{center}
    \scalebox{0.7}{
    \setlength{\tabcolsep}{0.8mm}{
   \begin{tabular}{|c|c|c|c|c|c|}
        \hline
        \multirow{2}{*}{Method}  & \multirow{2}{*}{Optical flow} & \multicolumn{2}{|c|}{UCF101} & \multicolumn{2}{|c|}{HMDB51} \\
        \cline{3-6}
         & & top-1 & top-5 & top-1 & top-5 \\
        \hline\hline
        Two-stream~\cite{simonyan2014two} & \checkmark & 86.9 & - & 58.0 & - \\
        Two-stream (+SVM)~\cite{simonyan2014two} & \checkmark & 88.0 & - & 59.0 & - \\
        I3D~\cite{i3d} & \checkmark & \textbf{98.0} & - & \textbf{80.7} & - \\
        \hline
        TSN~\cite{wang2016temporal} & $\times$ & 85.1 & - & 51.0 & - \\
        I3D-RGB~\cite{i3d} & $\times$& 84.5 & - & 49.8 & - \\
        TBN~\cite{li2019temporal} & $\times$ & 89.6 & - & 62.2 & - \\
        \hline
        Motion path & $\times$& 87.0 & 97.9 & 55.4 & 85.4 \\
        Our two-path & $\times$& \textbf{90.6} & \textbf{98.6} & 55.4 & \textbf{86.6} \\
        \hline
    \end{tabular}}}
    \end{center}
    \caption{Two-path results on UCF101 and HMDB51. Accuracies are calculated by averaging results from 3 splits. The size of the input clips for the state-of-the-art method I3D is $8\times$ our motion path input and the network parameters are around $2\times$ larger. Our two-path network is even better than the basic two-stream model which required optical flow features.}
    \label{table:twopath_ucf}
    \vspace{-10pt}
\end{table}

\begin{table}[tb]
    \begin{center}
    \scalebox{0.7}{
    \setlength{\tabcolsep}{2mm}{
   \begin{tabular}{|c|c|c|c|}
        \hline
        Method & Optical flow & top-1 & top-5 \\
        \hline\hline
        TBN C2D~\cite{li2019temporal} & $\times$ & 69.0 & 89.8 \\
        TBN C3D~\cite{li2019temporal} & $\times$ & 67.2 & 88.3 \\
        MARS~\cite{crasto2019mars} & $\times$ & 72.3 & - \\
        MARS + RGB~\cite{crasto2019mars} & $\times$ & 72.8 & -\\
        MARS + RGB + Flow ~\cite{crasto2019mars}& \checkmark & 73.5 & - \\
        \hline
        Motion path & $\times$ & 64.4 & 86.4 \\
        Our two-path & $\times$ & \textbf{73.9} & \textbf{91.4} \\
        \hline
    \end{tabular}}}
    \end{center}
    \caption{Results on Mini-kinetics. Our tow-path network outperforms MARS even when it is combined with another RGB and optical stream. The depth of our motion path is 18 while that for MARS is 101.}
    \label{table:twopath_mini}
    \vspace{-10pt}
\end{table}

Here, we do not focus on developing a new network architecture, and therefore, we only compare our method with some corresponding methods, as shown in Table~\ref{table:twopath_ucf}. Our single motion path can outperform TSN~\cite{wang2016temporal} and I3D-RGB~\cite{i3d} which only use RGB input data. Without any additional computation for optical flow and only using ResNet18, we can have better performance than the original two-stream model~\cite{simonyan2014two} which uses optical flow. On the other hand, our model is not better than the state-of-the-art such as~\cite{i3d}. But it is out of the scope of our paper because many settings including the input size and network architectures are totally different.

For Mini-kinetics, results are shown in Table~\ref{table:twopath_mini}. We mainly compared our method with TBN~\cite{li2019temporal} and MARS~\cite{crasto2019mars}, which does not use optical flow yet achieving good performances. WTBN used temporal bilinear modeling to process temporal information, which is insufficient to extract motion features compared with ours. The backbone network for MARS is ResNeXt-101-3D. To get the results using distillation methods, their networks should be trained on optical flow inputs first, and then another network is built to learn features from optical flow stream. The process is complex and is much more expensive than  our proposed two-path method. The backbone network for our motion path is ResNet-18-3D, which is shallower than that used in MARS. There is much room for our proposed solution to improve by using deeper networks and other feature fusion methods.

\section{Conclusion}
In this paper, we mainly focused on extracting motion features without optical flow. 3D ConvNets are believed to be capable of capturing motion features when RGB frames are set as input, but we demonstrated that it is not always true. We improved use of 3D convolution by using stacked residual frames as the network input. The overhead for this computation was negligibly small. With residual frames, the results of 3D ConvNets could be improved significantly when trained from scratch on UCF101 and HMDB51 datasets. Besides residual frames input, we proposed a two-path network, of using the motion path to extract motion features while the appearance path uses RGB frames to get the corresponding appearance. By combining the results from two paths, the state-of-the-art could be achieved on the Mini-kinetics dataset and better or comparable results can be achieved on UCF101 and HMDB51 datasets compared with the corresponding two-stream methods. Our results and analysis imply that residual frames can be a fast but effective way for a network to capture motion features and they are a good choice for avoiding complex computation for optical flow. In our future work, we will focus on performance improvement by investigating better combination method for our two-path network.


{\small
\bibliographystyle{ieee_fullname}
\bibliography{egbib}
}

\end{document}